\documentclass{article}

\PassOptionsToPackage{numbers, compress}{natbib}
\usepackage{colortbl}
\usepackage[dvipsnames]{xcolor}
\usepackage[final,main]{robotwin_2025}
\usepackage{marvosym}

\usepackage[utf8]{inputenc} 
\usepackage[utf8]{inputenc} 
\usepackage{newunicodechar}
\newunicodechar{，}{,}
\usepackage[T1]{fontenc}    
\usepackage{pifont}
\usepackage{url}            
\usepackage{booktabs}       
\usepackage{amsfonts}       
\usepackage{nicefrac}       
\usepackage{microtype}      
\usepackage{xcolor}         
\usepackage{listings}
\usepackage{graphicx}
\usepackage{array}
\usepackage{amsmath}
\usepackage{amssymb}
\usepackage{adjustbox}
\usepackage{tcolorbox}
\usepackage{multirow}
\usepackage{makecell} 
\usepackage{tabularx}

\usepackage{booktabs}
\usepackage{multicol}
\usepackage{array}
\usepackage{xcolor}
\usepackage{colortbl}
\usepackage{caption}

\definecolor{RoboTwincolor1}{HTML}{65a487}
\definecolor{RoboTwincolor2}{HTML}{68349a}

\usepackage[breaklinks=true,
            colorlinks,
            linkcolor = RoboTwincolor1,
            urlcolor  = RoboTwincolor1, 
            citecolor = teal,
            bookmarks = false]{hyperref}

\title{
      \textbf{Benchmarking Generalizable Bimanual Manipulation: \textcolor{RoboTwincolor1}{Robo}\textcolor{RoboTwincolor2}{Twin} Dual-Arm Collaboration Challenge at CVPR 2025 MEIS Workshop} 
}

\author{
    \\ \vspace{-4em}
    \\
    \textbf{Tianxing Chen}$^{1}\textsuperscript{*}$, \textbf{Kaixuan Wang}$^{1}\textsuperscript{*}$, \textbf{Zhaohui Yang}$^{2}\textsuperscript{*}$, \textbf{Yuhao Zhang}$^{2}\textsuperscript{*}$, \textbf{Zanxin Chen}$^{6}\textsuperscript{*}$,\\
    \textbf{Baijun Chen}$^{8}\textsuperscript{*}$, \textbf{Wanxi Dong}$^{12}\textsuperscript{*}$, \textbf{Ziyuan Liu}$^{5}$, \textbf{Dong Chen}$^{5}$, \textbf{Tianshuo Yang}$^{1}$, \\ 
    \textbf{Haibao Yu}$^{1}$, \textbf{Xiaokang Yang}$^{2}$, \textbf{Yusen Qin}$^{3}$, \textbf{Zhiqiang Xie}$^{4}$, \textbf{Yao Mu}$^{2 \text{\Letter}}$, \textbf{Ping Luo}$^{1 \text{\Letter}}$\\
    \textbf{and} \textbf{\textcolor{RoboTwincolor1}{All Competition} \textcolor{RoboTwincolor2}{Volunteers and Participants}}$^{1-27}$ (Full List can be found in Sec.~\ref{Competition Participants}) \\ \\
    $^1$HKU MMLab, $^2$SJTU, $^3$D-Robotics, $^4$AgileX Robotics, $^5$Huawei Germany, $^6$SZU, $^7$THU, $^{8}$NJU,\\$^{9}$VIVO, $^{10}$Jingdong Technology Information Technology Co., Ltd., $^{11}$Horizon Robotics, $^{12}$SUST,\\$^{13}$USST,$^{14}$SYSU, $^{15}$UESTC, $^{16}$SCU, $^{17}$Dexmal, $^{18}$SWJTU, $^{19}$HKUST, $^{20}$BJTU, $^{21}$BUAA,\\$^{22}$NEU, $^{23}$SHU, $^{24}$Sangfor Technologies Inc., $^{25}$CAUC, $^{26}$HUST, $^{27}$Reconova Technologies Co.\\
    $\textsuperscript{*}$ Equal contribution \quad $^{\text{\Letter}}$ Corresponding authors \\ \\
    \textbf{\href{https://robotwin-benchmark.github.io/cvpr-2025-challenge/}{\textcolor{RoboTwincolor1}{https://robotwin-benchmark.github.io/}\textcolor{RoboTwincolor2}{cvpr-2025-challenge/}}}
}

\begin{document}

\maketitle

\vspace{-2em}
\begin{abstract}
Embodied Artificial Intelligence (Embodied AI) is an emerging frontier in robotics, driven by the need for autonomous systems that can perceive, reason, and act in complex physical environments. While single-arm systems have shown strong task performance, collaborative dual-arm systems are essential for handling more intricate tasks involving rigid, deformable, and tactile-sensitive objects. To advance this goal, we launched the \textbf{\textcolor{RoboTwincolor1}{Robo}\textcolor{RoboTwincolor2}{Twin} Dual-Arm Collaboration Challenge} at the 2nd MEIS Workshop, CVPR 2025. Built on the RoboTwin Simulation platform (1.0~\cite{mu2025robotwin,mu2024robotwin}, 2.0~\cite{chen2025robotwin}) and the AgileX COBOT-Magic Robot platform, the competition consisted of three stages: \textit{Simulation Round 1}, \textit{Simulation Round 2}, and a final \textit{Real-World Round}. Participants totally tackled 17 dual-arm manipulation tasks, covering rigid, deformable, and tactile-based scenarios. The challenge attracted 64 global teams and over 400 participants, producing top-performing solutions like SEM~\cite{lin2025sem} and AnchorDP3~\cite{2506.19269} and generating valuable insights into generalizable bimanual policy learning. This report outlines the competition setup, task design, evaluation methodology, key findings and future direction, aiming to support future research on robust and generalizable bimanual manipulation policies.
\end{abstract}
\section{Introduction}
Embodied Artificial Intelligence (Embodied AI~\cite{embodiedaiguide2025}) has emerged as a critical frontier in modern robotics, driven by the growing demand for autonomous agents capable of seamlessly perceiving, reasoning about, and interacting with the physical world. While single-agent systems have made significant strides in recent years, the next frontier lies in the development of collaborative multi-agent systems and precise, adaptive manipulation capabilities that can operate across a wide variety of environments and task complexities.

Key challenges in this domain include long-horizon manipulation of rigid objects, coordinated bimanual interactions, and increasingly, the deformable object manipulation necessary for handling flexible materials such as cloth, towels, and cables—tasks that are particularly challenging due to their high-dimensional and underactuated dynamics. Moreover, the integration of tactile sensing modalities introduces additional potential for fine-grained control and real-world generalization, yet presents new algorithmic difficulties in tactile-conditioned policy learning and multimodal fusion.

This rapidly evolving landscape has led to a wide spectrum of technical approaches, ranging from different input modalities (2D RGB, 3D geometry, RGB-D hybrid 2.5D, and tactile signals), to architectural paradigms spanning classical perception-action pipelines, learning-based frameworks, and more recently, large multimodal transformer systems capable of unifying language, vision, and low-level control.

Despite these advancements, the field still grapples with several fundamental questions: (1) How can agents learn robust and transferable skills from noisy, real-world multimodal data? (2) How can we effectively balance data composition in multi-task policy learning? (3) How can models generalize across unseen objects, environments, and embodiments? (4) How can tactile and visual modalities be fused in a way that reflects their complementary strengths? (5) How do we ensure sample efficiency, scalability, and real-time deployment feasibility, especially under limited computational resources, using techniques like quantization, distillation, or policy modularization?

To tackle these challenges, we introduce the \textbf{\textcolor{RoboTwincolor1}{Robo}\textcolor{RoboTwincolor2}{Twin} Dual-Arm Collaboration Challenge}, held as part of the 2nd MEIS Workshop at CVPR 2025. This competition invites the global research community to explore the frontiers of dual-arm manipulation across both simulation and real-world settings. Leveraging the RoboTwin simulation benchmark and the AgileX Cobot-Magic physical platform, participants are encouraged to develop generalizable, robust, and multimodal policies capable of solving diverse manipulation tasks involving rigid objects, deformable materials, and tactile-informed interactions.

Through this challenge, we aim to not only benchmark current capabilities but also push the boundary of embodied intelligence, setting new standards for collaborative, multimodal robot agents in complex physical environments.

\section{Competition Structure, Round-wise Rules and Results Overview}
\subsection{Simulation Round 1}

\subsubsection{Round-wise Rules}

\begin{figure}[h] 
    \vspace{-10pt}
    \centering    \includegraphics[width=1.0\linewidth]{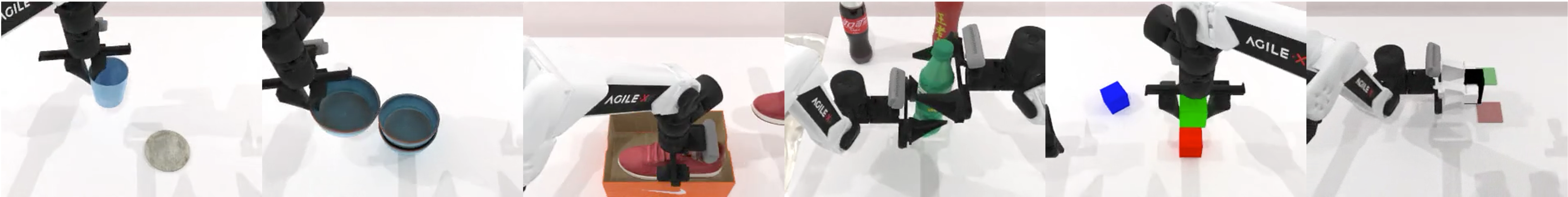}
    \vspace{-15pt}
    \caption{\textbf{Simulation Round 1 Tasks (5 Rigid Object Manipulation Tasks and 1 Tactile Manipulation Task).}}
    \label{fig:round-1-tasks} 
    \vspace{-1em}
\end{figure}

To evaluate the current capabilities of manipulation policies and the technical maturity of participating researchers, we designed five rigid-body dual-arm manipulation tasks and one novel visuo-tactile object sorting task on the RoboTwin 1.0~\cite{mu2025robotwin,mu2024robotwin} simulation platform (\textit{Place Empty Cup}, \textit{Stack Bowls Three}, \textit{Put Dual Shoes}, \textit{Put Bottles Dustbin}, \textit{Stack Blocks Three} and \textit{Classify Tactile}). The visuo-tactile task incorporates object deformation simulated via finite element analysis (FEA), presenting a unique challenge that requires multimodal reasoning~\cite{li2024maniskill}. Key frames of the six tasks are shown in Fig.~\ref{fig:round-1-tasks}.

Each task allows the submission of a single dedicated model, with a maximum score of 20 points per task (5 points for the tactile task), leading to a total score of 105 points. Notably, stage-wise scoring was employed to provide finer-grained differentiation among participants' performance and ensure better resolution in ranking. Participants were permitted to collect an unlimited amount of data using the RoboTwin platform during the development phase. To ensure fair evaluation, all test-time seeds—corresponding to randomized scenes and object configurations—were kept unseen during training. Furthermore, the final inference models were required to run on a single RTX 4090 GPU, enforcing a standardized computational constraint.

To reduce randomness and ensure reliability, each task was evaluated over 100 independent trials. Notably, environmental attributes such as background texture, table and wall colors, lighting conditions, and table height were kept consistent between training and evaluation environments. All Task Details can be found at \href{https://robotwin-benchmark.github.io/cvpr-2025-challenge/}{https://robotwin-benchmark.github.io/cvpr-2025-challenge/}.

\subsubsection{Results Overview}

We present the average competition performance per task in Round 1 under two evaluation settings: valid submissions only and all submissions (with non-submissions counted as zero), as shown in the Tab.~\ref{tab:round1-overview-table-transposed}. It can be observed that participants achieved impressive average success rates across many tasks, with the highest success rate for all tasks exceeding 97. This indicates that the tasks in Round 1 were relatively simple and well within the capabilities of existing algorithms, primarily because the evaluation scenes were aligned with the training environments.

\begin{figure*}[h]
    \centering
    \footnotesize
    \captionof{table}{\textbf{Overall Per-task Performance in Simulation Round 1.}}
    \vspace{2pt}
  \resizebox{0.9\textwidth}{!}{%
    \begin{tabular}{lccc}
      \toprule
      \textbf{Task} & \textbf{Average (Valid submissions only)} & \textbf{Average (All submissions)} & \textbf{Max} \\
      \midrule
      \textit{Place Empty Cup} & 84.1 & 75.1 & 100.0 \\
      \textit{Stack Bowls Three} & 79.6 & 62.5 & 99.0 \\
      \textit{Place Dual Shoes} & 57.4 & 45.1 & 99.3 \\
      \textit{Put Bottles Dustbin} & 76.8 & 63.1 & 97.0 \\
      \textit{Stack Blocks Three} & 51.9 & 40.8 & 100.0 \\
      \textit{Classify Tactile} & 73.0 & 26.1 & 100.0 \\
      \bottomrule
    \end{tabular}
    }
    \label{tab:round1-overview-table-transposed}
\end{figure*}

\begin{figure*}[h]
  \centering
  \begin{minipage}{0.43\textwidth}
    \centering
    \includegraphics[width=1.0\linewidth]{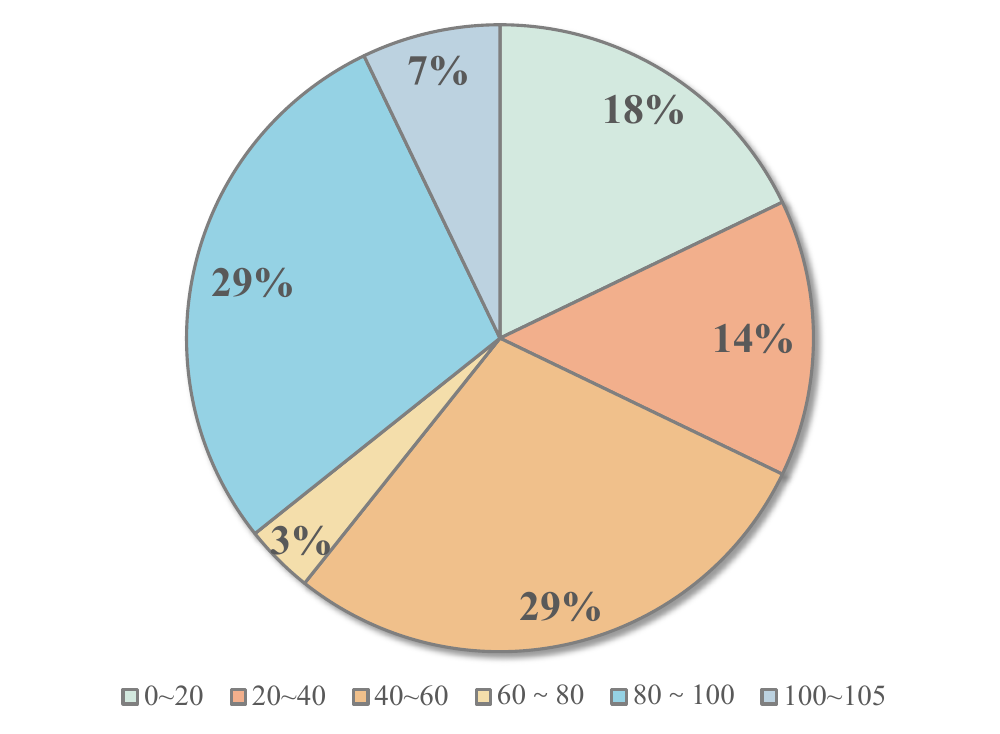}
    \vspace{-1.5em}
    \captionof{figure}{\textbf{Distribution of Team Scores in Round 1.}}
    \label{fig:round-1-dist} 
  \end{minipage}
  \hfill
  \begin{minipage}{0.56\textwidth}
    \centering
    \includegraphics[width=1.0\linewidth]{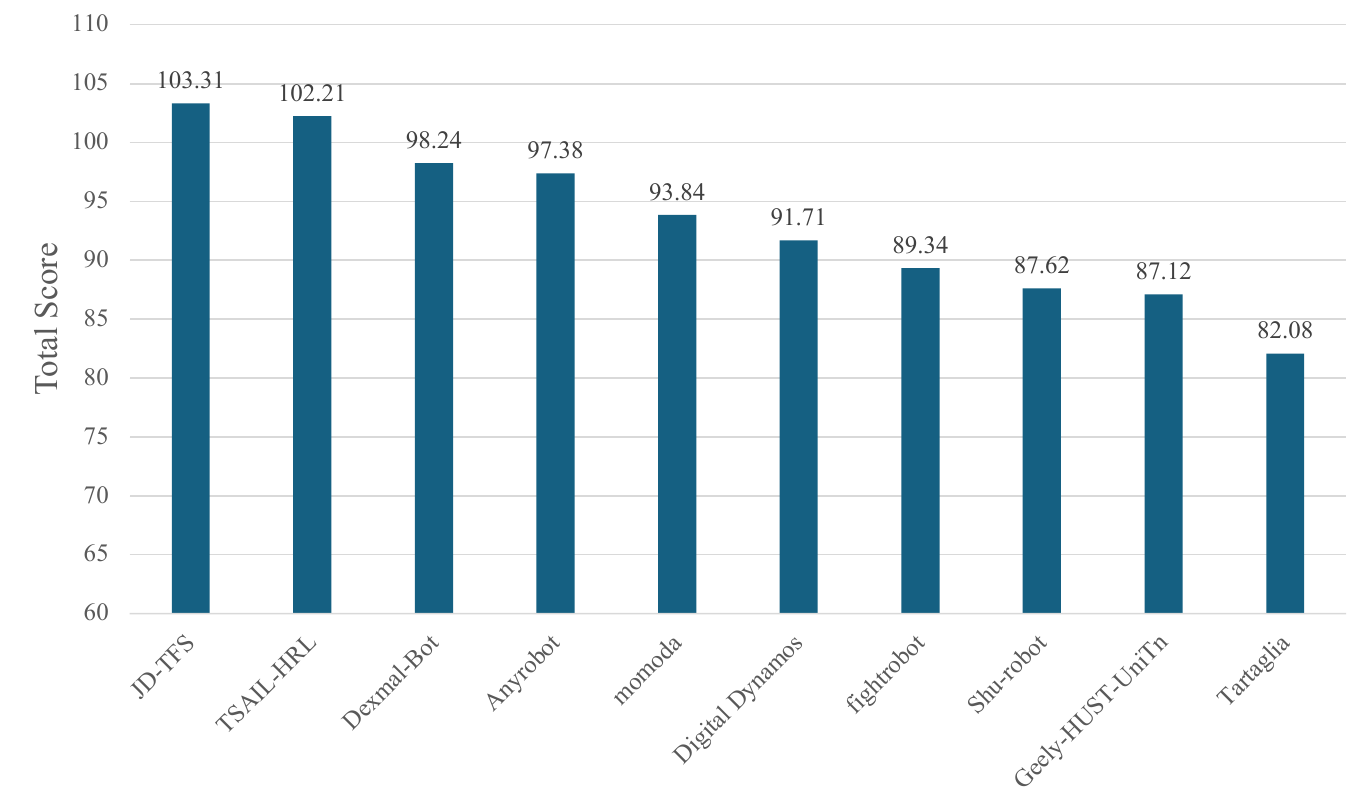}
    \vspace{-1.5em}
    \captionof{figure}{\textbf{Top 10 Team Scores in Round 1.}}
    \label{fig:round-1-top10} 
  \end{minipage}
\end{figure*}

From the distribution, most teams scored above 60 in total (Fig.~\ref{fig:round-1-dist}), and the performance of the top 10 teams exhibits an approximately linear progression, as shown in Fig.~\ref{fig:round-1-top10}. This indicates that the overall difficulty of the challenge was well-balanced and appropriately calibrated.

\subsection{Simulation Round 2}

\subsubsection{Round-wise Rules}

\begin{figure}[h] 
    \centering    \includegraphics[width=1.0\linewidth]{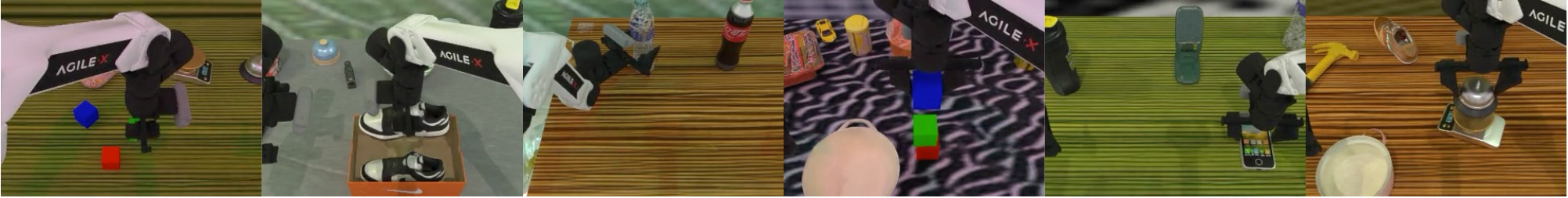}
    \vspace{-15pt}
    \caption{\textbf{Simulation Round 2 Tasks (6 Tasks with domain randomization).}}
    \label{fig:round-2-tasks} 
    \vspace{-1em}
\end{figure}

To further challenge the limits of current policy generalization and uncover more advanced solutions for dual-arm manipulation, Round 2 of the competition adopted the enhanced RoboTwin 2.0~\cite{chen2025robotwin} platform. Compared to Round 1, we increased the difficulty across multiple dimensions, including visual robustness, multi-task handling, and policy adaptability. Six rigid-body manipulation tasks were introduced (\textit{Blocks Ranking RGB}, \textit{Place Dual Shoes}, \textit{Put Bottles Dustbin}, \textit{Stack Blocks Three}, \textit{Place Phone Stand}, \textit{Place Object Scale}), as shown in Fig.~\ref{fig:round-2-tasks}, and participants were required to develop a single unified model capable of solving all tasks. To support task disambiguation, language instructions—unseen during evaluation—were provided, requiring the model to interpret natural language prompts to execute the correct behavior. During evaluation, we introduced domain randomization including diverse and unseen background textures, scene clutter, ±3 cm variations in table height, and changing lighting conditions to test the model’s robustness. Importantly, tasks had to be fully completed to receive any score, with each task scored out of 100 for a total of 600 points. All Task Details can be found at \href{https://robotwin-benchmark.github.io/cvpr-2025-challenge/}{https://robotwin-benchmark.github.io/cvpr-2025-challenge/}.

\subsubsection{Results Overview}

\begin{figure*}[h]
  \centering
  \begin{minipage}{0.45\textwidth}
     \centering
    \footnotesize
    \captionof{table}{\textbf{Overall Per-task Performance in Simulation Round 2.}}
  \resizebox{0.95\textwidth}{!}{
    \begin{tabular}{lccc}
      \toprule
      \textbf{Task} & \textbf{Average} & \textbf{Max} \\
      \midrule
      \textit{Blocks Ranking RGB} & 42.4 & 100.0 \\
      \textit{Place Dual Shoes}  & 45.0 & 98.0 \\
      \textit{Put Bottles Dustbin} & 51.8 & 100.0 \\
      \textit{Stack Blocks Three}  & 50.2 & 100.0 \\
      \textit{Place Phone Stand} & 43.8 & 100.0 \\
      \textit{Place Object Scale} & 32.8 & 100.0 \\
      \bottomrule
    \end{tabular}
    }
    \label{tab:round2-overview-table-transposed}
  \end{minipage}
  \hfill
  \begin{minipage}{0.54\textwidth}
    \centering
    \includegraphics[width=1.0\linewidth]{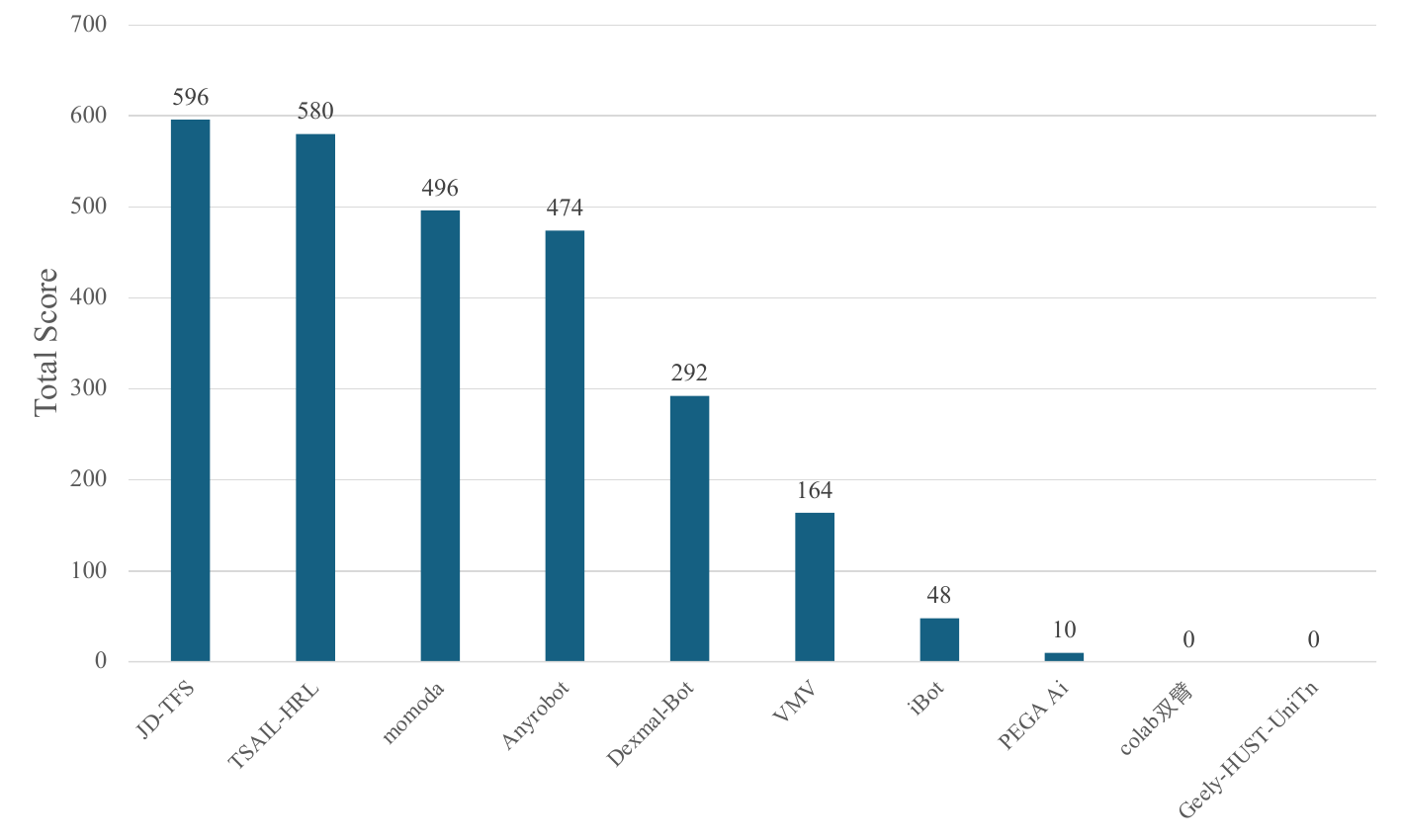}
    \vspace{-1.8em}
    \captionof{figure}{\textbf{Top 10 Team Scores in Round 2.}}
    \label{fig:diverse-grasp} 
  \end{minipage}
\end{figure*}

We present the average scores per task and top-10 team performances from Simulation Round 2. Each task is scored out of 100 points, with a total possible score of 600. As task difficulty increased, we observed a wider performance gap among the top teams, alongside a noticeable drop in the overall average score. This indicates that after applying domain randomization, many conventional strategies struggled to maintain robustness under the more challenging conditions. Nevertheless, a few top-performing teams were able to achieve near-perfect scores through carefully designed architectures and learning schemes. For further details, please refer to Sec.~\ref{AnchorDP3}.

\subsection{Real-World Round}

\subsubsection{Round-wise Rules}

\begin{figure}[h] 
    \centering    \includegraphics[width=1.0\linewidth]{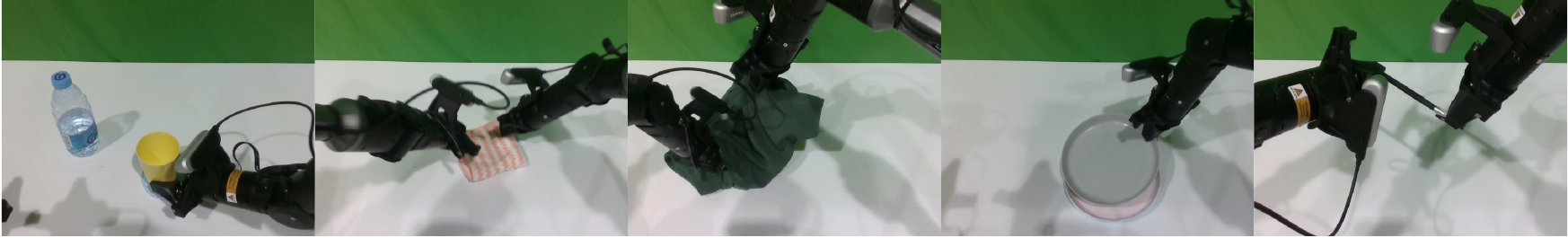}
    \vspace{-15pt}
    \caption{\textbf{Real-World Round Tasks.}}
    \label{fig:round-real-tasks} 
    \vspace{-5pt}
\end{figure}

We designed Five dual-arm manipulation tasks based on the COBOT-Magic platform, including \textit{Pour Water}, \textit{Fold Towel}, \textit{Fold Shorts}, \textit{Stack Plates}, and \textit{Cap Pen}, as shown in Fig.~\ref{fig:round-real-tasks}. To simulate real-world variances during large-scale data collection, we initially provide 300 demonstrations per task collected in non-competition environments. These demonstrations feature slight variations in camera viewpoints, table height, background clutter, and diverse tabletop textures. Participants are expected to leverage this variability to extract transferable knowledge.

One week prior to the final code and model submission deadline, we released an additional 20 high-quality demonstrations per task. These were collected in the official competition setting, featuring a clean tabletop and fixed, visible backgrounds, and were intended to represent the target evaluation domain. Notably, during final evaluation, 15 out of 20 trials per task were conducted under this seen configuration, while the remaining 5 trials used unseen background variations to assess visual generalization and robustness.

\begin{figure}[h] 
    \vspace{-5pt}
    \centering    \includegraphics[width=1.0\linewidth]{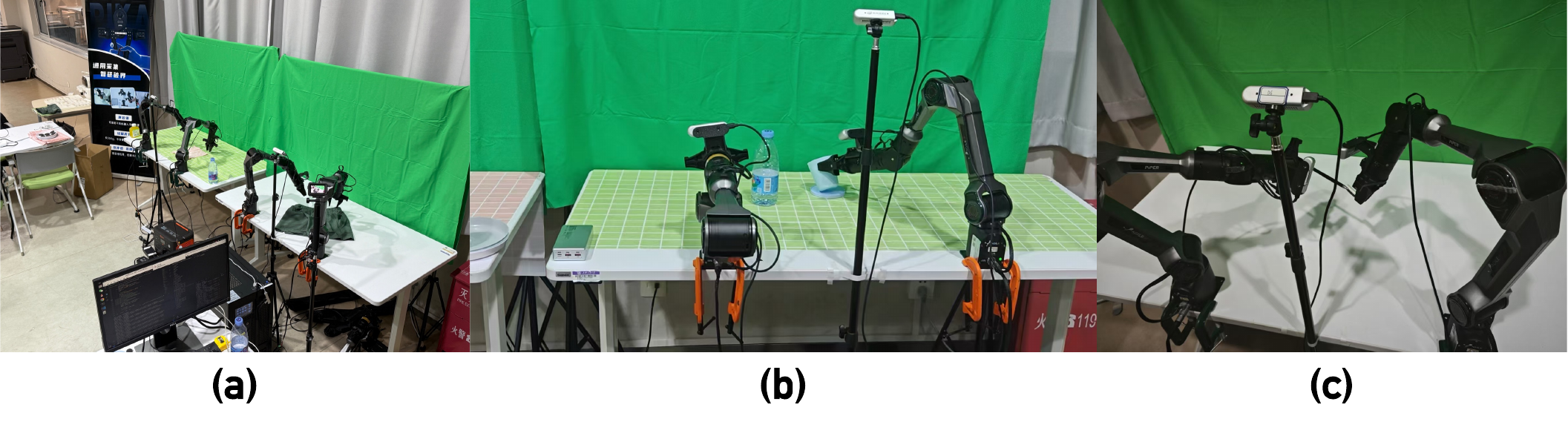}
    \vspace{-20pt}
    \caption{\textbf{Real-World Challenge Demonstration.}}
    \label{fig:real-world-demo} 
\end{figure}

Each task is evaluated over 20 trials. The first 15 trials are conducted under the clean, seen tabletop setting, while the final 5 trials are evaluated with unseen background cloths, to test the model’s visual robustness. Importantly, participants must develop a single model and shared set of weights to solve all six tasks. To prevent hard-coded solutions, language instructions used during evaluation will not be provided during training. The demonstration of real-world challenge is shown in Fig.~\ref{fig:real-world-demo}. Subfigure (a) shows the real-world evaluation setup, where two dual-arm Piper robots were deployed in identical physical environments with consistent table settings and background cloths. Subfigures (b) and (c) illustrate representative strategy executions for the tasks Pour Water and Cap Pen, respectively, as performed by participating teams.

\subsubsection{Results Overview}

\begin{figure*}[h]
  \centering
  \begin{minipage}{0.45\textwidth}
     \centering
    \footnotesize
    \captionof{table}{\textbf{Overall Per-task Performance in Real World Round.}}
    \vspace{2pt}
  \resizebox{0.75\textwidth}{!}{%
    \begin{tabular}{lccc}
      \toprule
      \textbf{Task} & \textbf{Average} & \textbf{Max} \\
      \midrule
      \textit{Pour Water} & 1.31 & 6.0 \\
      \textit{Fold Towel}  & 0.30 & 2.1 \\
      \textit{Fold Shorts} & 0.59 & 3.8 \\
      \textit{Stack Plates}  & 6.80 & 17.0 \\
      \textit{Cap Pen} & 0.58 & 3.1 \\
      \bottomrule
    \end{tabular}
    }
    \label{tab:real-overview-table-transposed}
  \end{minipage}
  \hfill
  \begin{minipage}{0.54\textwidth}
    \centering
    \includegraphics[width=0.9\linewidth]{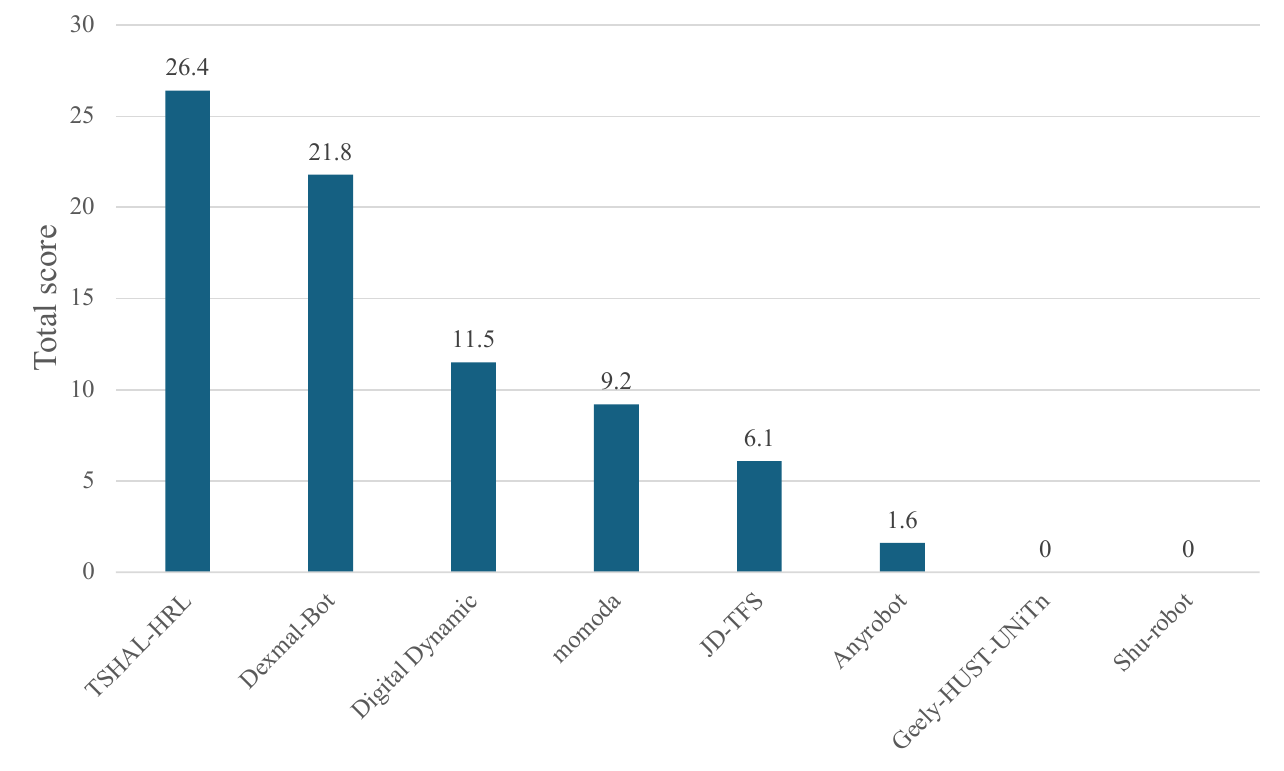}
    \vspace{-1.2em}
    \captionof{figure}{\textbf{Top 10 Team Scores in Real-World Round.}}
    \label{fig:diverse-grasp} 
  \end{minipage}
\end{figure*}

We present the per-task average scores and the overall scores of the top 8 teams in the Real-World Track. Each task is scored out of 20 points, with a total possible score of 100. The results show that real-world dual-arm manipulation remains highly challenging—particularly due to factors such as deformable object interactions and long-horizon task dependencies. Many learned strategies failed to generalize effectively, and even the best-performing team achieved only 26.4 points. This underscores the substantial research opportunities that remain in developing reliable dual-arm policies for complex real-world scenarios.

\section{Excellent Solutions}


We highlight the winning solutions, AnchorDP3~\cite{2506.19269} by the JD-TFS Team and SEM~\cite{lin2025sem} by the TSAIL-HRL Team. Interestingly, both approaches incorporate explicit 3D modalities, in contrast to many recent VLA-based methods that rely solely on multi-view 2D visual inputs. This suggests that leveraging structured 3D representations can significantly enhance policy learning by improving sample efficiency and providing a more grounded understanding of physical space.

\subsection{AnchorDP3 by JD-TFS Team}
\label{AnchorDP3}

\begin{figure*}[h]
    \centering
    \includegraphics[width=1.0\linewidth]{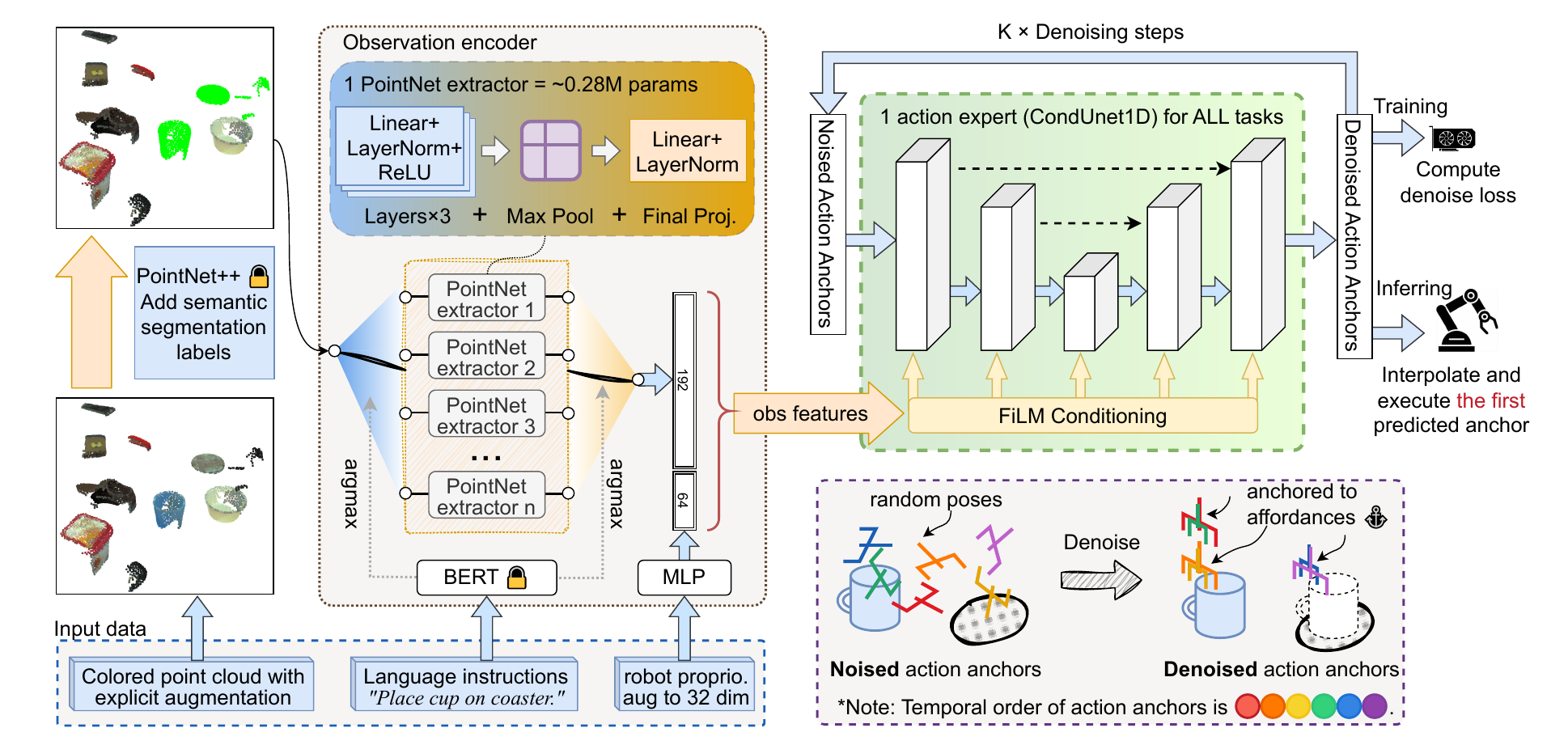}
    \captionof{figure}{\textbf{Pipeline of AnchorDP3.}}
    \label{fig:sim-champ-pipeline} 
\end{figure*}

The JD-TFS team's AnchorDP3~\cite{2506.19269} framework secured the championship in the Simulation Track of the Challenge, including both Simulation Round 1 and Round 2. This achievement was enabled by a fundamental rethinking of how robotic manipulation policies should be learned. AnchorDP3 achieved a remarkable 98.7\% success rate through a paradigm shift in action representation and the introduction of three complementary innovations that directly tackle the core challenges of dual-arm manipulation in highly randomized environments. The AnchorDP3 pipeline is shown in Fig.~\ref{fig:sim-champ-pipeline}.

The team's most significant insight was recognizing that conventional diffusion policies waste computational resources on predicting dense action sequences (20-25Hz), where most actions are trivial, momentum-driven movements. Instead, AnchorDP3 predicts only geometrically meaningful "keyposes" - critical transition points anchored to object affordances such as pre-grasp, grasp, and placement positions. This mirrors human motor control, where conscious planning occurs only at kinematic inflection points while transit phases remain subconscious.

This sparse representation offers multiple advantages: (1) Reduced prediction space: Instead of predicting thousands of dense actions, the policy learns only 10-30 sparse keyposes per task (2) Better causality learning: The policy learns affordance-driven decisions ("move toward the object") rather than spurious correlations ("move forward because I was moving forward") (3) 14.5x more diverse training data: Within the same computational budget, the team could expose their model to vastly more environmental variations.

The team leveraged the simulation environment's complete scene knowledge to automatically generate precise point-level segmentation masks for task-critical objects. This eliminated perceptual ambiguity in cluttered scenes without manual annotation.

Rather than forcing a single encoder to handle all tasks, they employed lightweight, task-specific encoders (only ~0.28M parameters each) that feed into a shared diffusion action expert. This modular design prevented negative interference between different manipulation strategies while maintaining computational efficiency.

The policy simultaneously predicts both joint angles and end-effector poses, exploiting geometric consistency to accelerate convergence and improve accuracy. Although only the first predicted keypose is executed, supervising all predicted keyposes and their full kinematic states significantly enhanced learning stability.

The team addressed fundamental inefficiencies in existing approaches rather than merely scaling up computation. By reformulating the action space around sparse, meaningful decisions, they achieved superior performance with less data per trajectory while dramatically increasing environmental diversity exposure. The combination of efficient representation learning, modular architecture, and clever use of simulation capabilities created a solution that was both theoretically principled and practically effective.

\subsection{SEM by TSAIL-HRL Team}

\begin{figure*}[h]
    \centering
    \includegraphics[width=0.98\linewidth]{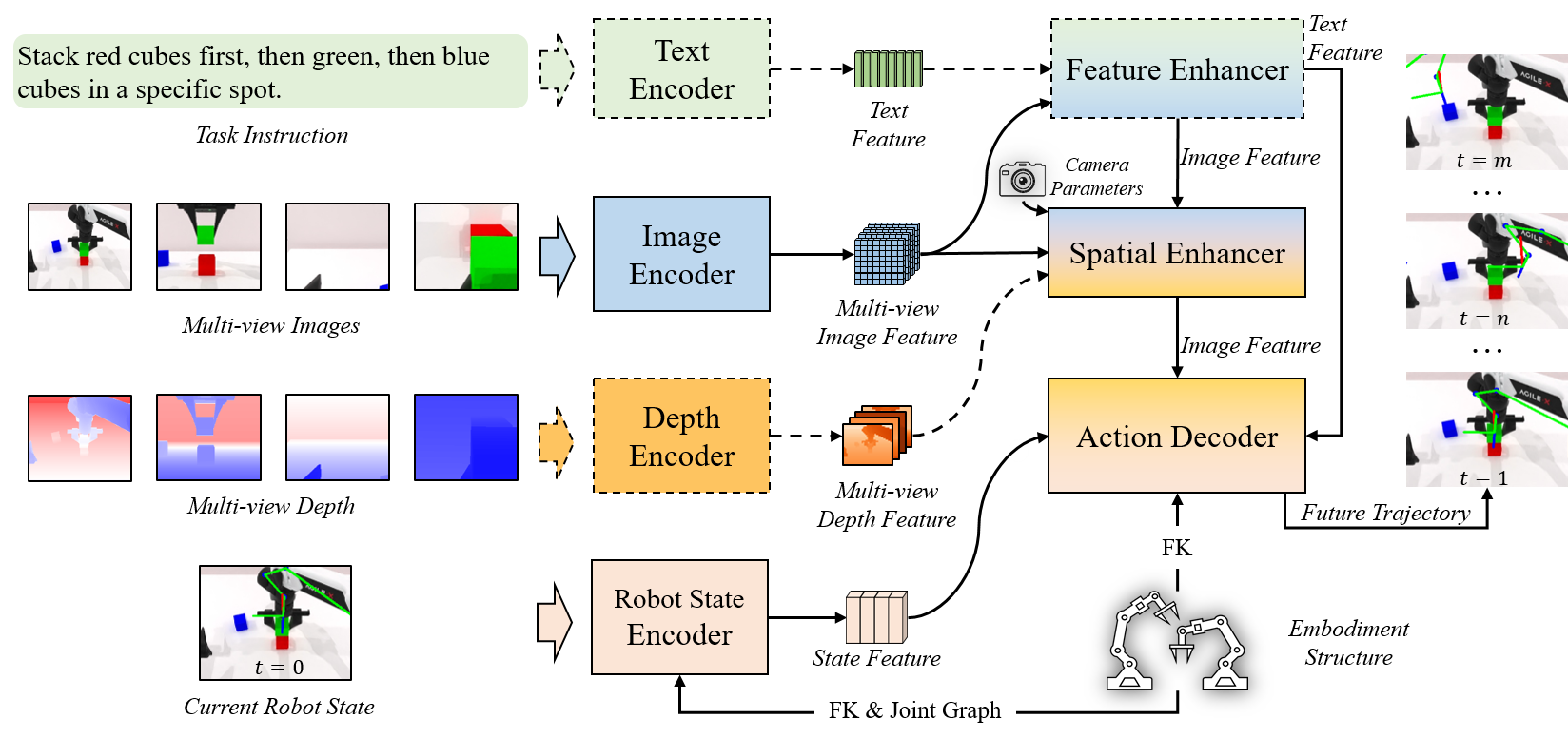}
    \captionof{figure}{\textbf{Pipeline of SEM.}}
    \label{fig:real-world-champ-pipeline} 
\end{figure*}

The TSAIL-HRL team's SEM~\cite{lin2025sem} framework demonstrated remarkable performance in the simulation rounds, which reconceptualizes robot manipulation as a joint problem of explicit 3D spatial reasoning and embodiment-aware control. Rather than treating perception and action as loosely coupled subproblems, SEM tightly integrates multi-view vision, depth cues, and full kinematic state into a unified diffusion policy. At the architectural level, SEM comprises a spatial enhancer that lifts 2D image features into a coherent 3D embedding space, a robot state encoder that captures the full joint-graph structure of the dual-arm system, and a diffusion-based action decoder that conditions on fused semantic and geometric representations. The SEM pipeline is shown in Fig.~\ref{fig:real-world-champ-pipeline}.

This design is motivated by two core insights. First, purely 2D visual encoders struggle with depth ambiguity in cluttered scenes—a limitation overcome by sampling candidate depths across views and projecting pixel features into 3D positional embeddings, thereby preserving 2D semantics while enabling precise spatial understanding. Second, end-effector–only representations discard rich proprioceptive information; by modeling each joint as a node in a graph and applying attention over inter-joint distances, the robot state encoder yields an embodiment-aware embedding that respects the system’s physical structure. Crucially, SEM’s plug-and-play modules prevent negative transfer across tasks and allow independent upgrading of vision, language, or control components without retraining the entire policy.

Empirically, SEM demonstrates that these design choices translate into both robustness and efficiency. The spatial enhancer and robot state encoder each contribute significant gains in ablation studies. Despite its theoretical rigor, SEM remains computationally tractable: lightweight, task-specific encoders minimize overhead, while the diffusion decoder provides a unified decision process. The result is a manipulation policy that generalizes gracefully to novel object arrangements, maintains high accuracy in the face of perceptual noise, and achieves superior task performance with fewer demonstrations and reduced training time.
\section{Key Insights}

The RoboTwin Challenge provided a rigorous testbed for evaluating data-driven dual-arm manipulation policies under both simulated and real-world settings. Participating teams explored diverse combinations of model architectures, data collection pipelines, and training strategies. Through a careful analysis of their solutions and outcomes, we distill several cross-cutting insights that are critical for advancing the field of embodied robotic manipulation. These insights are summarized below by thematic area.

\subsection{Aligning Model Capacity with Task Complexity}

Selecting model architectures that match the inherent complexity of the task proved essential for success. While simple tasks could be handled by lightweight policies, more intricate manipulations—particularly those requiring long-horizon planning or multi-object coordination—demanded higher-capacity models. The MOMODA team exemplified this trade-off: their success rate on the dual-shoes placement task rose from 48.2\% to 95.1\% after switching from a lightweight model to the VLA-based $\pi_0$ and scaling training episodes from 100 to 3000. This highlights the importance of aligning model expressiveness with task difficulty.

\subsection{Importance of Data Quantity and Quality}

Scaling up training data volume consistently led to improved performance across all tasks. Teams that transitioned from hundreds to thousands of training episodes benefited from increased environmental and interaction diversity. However, data \textit{quality} emerged as equally vital: high-fidelity demonstrations released shortly before the final round played a pivotal role in model fine-tuning and adaptation to real-world domains. The JD-TFS team adopted a two-stage training process—initial adaptation using large-scale medium-quality data (LSMQ-D), followed by fine-tuning with small-scale high-quality data (SSHQ-D)—which proved effective in bridging both embodiment and domain gaps. This underscores that scaling data quantity must be accompanied by robust quality assurance practices.

\subsection{Multi-Modal Fusion and Depth-Aware Encoding}

Successful policies effectively fused visual, depth, and language modalities to improve perception and generalization. The MOMODA team employed a two-stage depth integration strategy: first aligning a frozen depth encoder with existing visual embeddings, then jointly optimizing all components. This led to significant gains in policy accuracy. In addition, temporal downsampling—reducing the action frequency from 25Hz to 12Hz—improved the ability to predict long-horizon behaviors while maintaining computational efficiency. Such multi-modal integration is particularly important in embodied settings, where single-modality inputs are often insufficient for resolving complex manipulation dynamics.

\subsection{Instruction Grounding and Language Robustness}

Instruction-conditioned manipulation introduced new challenges in semantic grounding and task ambiguity. Notably, the VMV team found that simple, high-level instructions often outperformed complex, overly detailed prompts—suggesting that succinct language may generalize better across tasks. Several teams increased model robustness by augmenting training datasets with diverse instruction variants, helping mitigate the impact of noisy or ambiguous input prompts during evaluation. These findings call for improved alignment between linguistic intent and action affordances in embodied systems.

\subsection{Data Preprocessing and Demonstration Refinement}

In the real-world round, noisy demonstrations were a common issue, with problems such as inconsistent initial states, repeated grasp attempts, arm jitter, or irrelevant motions degrading model performance. The SHU-Robot team tackled this by designing a hybrid trajectory filtering pipeline combining automatic heuristics (e.g., start-frame detection, motion segmentation) with manual verification. They also applied temporal trimming and standardized trajectory lengths to ensure uniformity. This preprocessing step provided cleaner supervision signals, significantly boosting policy robustness. Such pipelines are likely to be essential components of future large-scale real-world datasets.

\subsection{Unified Model Generalization and Evaluation Bias}

The challenge requirement to deploy a single model for all tasks emphasized the difficulty of balancing learning across heterogeneous scenarios—ranging from long-horizon planning to deformable object manipulation. A telling example came from the Cap Pen task in the real-world round: the TSAIL-HRL team consistently executed most of the insertion sequence accurately but suffered from minor final-stage inaccuracies, leading to a low overall score of 3.1. This revealed a key shortcoming of outcome-focused evaluation metrics, which may under-reward policies demonstrating strong partial progress. Future benchmarks may benefit from progress-aware scoring schemes that reflect nuanced task achievement beyond binary success/failure.

\section{Competition Impact}

The RoboTwin Dual-Arm Collaboration Challenge was held as a sub-competition of the 2nd MEIS Workshop at CVPR 2025. It was jointly organized by leading institutions including The University of Hong Kong (MMLab@HKU), Huawei Germany, Shanghai Jiao Tong University, D-Robotics, AgileX Robotics, Stanford University, Imperial College London, University of Tennessee, and University of North Carolina at Chapel Hill.

The challenge attracted a total of 64 teams and over 400 participants from globally recognized institutions and organizations such as Tsinghua University, Peking University, Horizon Robotics, Moscow Institute of Physics and Technology, Waseda University, and City University of Hong Kong, among others.

The competition was widely promoted across multiple platforms, reaching an audience of over 100,000, with the official challenge website receiving more than 10,000 page views.
\section{Related Work}
\subsection{Benchmarks for Robotic Manipulation}

Recent years have seen a surge in benchmarks designed to advance robotic manipulation, spanning both simulation and real-world settings. On the simulation side, platforms such as SAPIEN and PartNet-Mobility enable dynamic interaction with a wide range of articulated objects, while ManiSkill2 offers millions of demonstration frames across diverse task families. Meta-World provides standardized tasks for meta- and multi-task learning, and CALVIN emphasizes long-horizon, language-conditioned tasks paired with rich sensory inputs. LIBERO focuses on continual skill acquisition with high-quality teleoperated demonstrations.

In parallel, large-scale real-world datasets have emerged to tackle the sim-to-real gap. AgiBot World delivers over one million human-verified trajectories across 217 real-world tasks. RoboMIND includes 107K teleoperated episodes with failure annotations spanning 479 tasks on four platforms. Open X-Embodiment consolidates demonstrations from 22 robot embodiments into a unified format for generalist policy learning, while Bridge offers 60K+ trajectories on low-cost hardware under diverse conditions. In contrast to these efforts, RoboTwin 1.0~\cite{mu2025robotwin,mu2024robotwin} introduced a bidirectional twin framework that pairs real teleoperated demonstrations with simulated replicas for domain-aligned evaluation. Building on this, RoboTwin 2.0~\cite{chen2025robotwin} integrates interactive LLM-driven feedback, applies domain randomization across visual, physical, and task parameters, and expands to include deformable object tasks and tactile-informed policy learning—positioning itself as a versatile benchmark for generalizable, multimodal dual-arm manipulation in both simulation and the real world.


\subsection{Robot Learning in Manipulation}

Recent advances in robotic manipulation have led to a wide range of policy architectures tailored for specific tasks and embodiments. Many task-specific approaches~\cite{ke20243d, ze20243d, chi2023diffusion, fu2024mobile, chen2025g3flow, dexhanddiff, wang2024rise, adaptdiffuser, liang2024skilldiffuser, 10900471, wen2025dexvla, lu2024manicm, liu2025avr} achieve strong single-task performance but struggle to transfer across different embodiments or generalize to novel task configurations.

In contrast, foundation models trained on million-scale, multi-robot corpora have enabled robust zero-shot generalization. RT-1~\cite{brohan2022rt-1} unifies vision, language, and action in a single transformer for real-time kitchen tasks; RT-2~\cite{brohan2023rt-2} co-fine-tunes large vision–language models on both web and robot data to unlock semantic planning and object reasoning capabilities. Diffusion-based models such as RDT-1B~\cite{liu2024rdt1b} and $\pi_0$~\cite{black2024pi_0} capture diverse bimanual dynamics from over a million episodes.

Vision–Language–Action (VLA) frameworks such as OpenVLA~\cite{openvla} and CogACT~\cite{li2024cogact}, along with adaptations like Octo~\cite{team2024octo}, LAPA~\cite{lapa}, and OpenVLA-OFT~\cite{openvla_oft}, further demonstrate the ability to efficiently fine-tune policies across new robots and sensing modalities, highlighting the growing potential for general-purpose embodied intelligence.

\section{Future Direction}


Building upon the results and observations from the RoboTwin Challenge, we identify several promising directions for advancing research in bimanual robotic manipulation:

(1) Long-Horizon and Multi-Stage Task Learning: Future efforts should focus on enabling agents to perform temporally extended tasks that require planning, memory, and coordination over multiple stages.

(2) Instruction-Following Manipulation: Incorporating natural language understanding into manipulation policies can unlock more flexible and intuitive human-robot interaction, especially for open-ended or zero-shot instructions.

(3) Deformable Object Manipulation: More sophisticated simulation, data generation, and policy architectures are needed to handle the complex dynamics and representations involved in manipulating deformable objects such as fabrics, towels, and flexible packaging.

(4) Self-Correction and Recovery Capabilities: Learning-based systems should develop mechanisms to detect and recover from errors autonomously, enabling higher task success rates in long-horizon settings.

(5) Robustness to Sensory Noise and Domain Shifts: Enhancing visual and multimodal perception to deal with observation noise, unseen backgrounds, and embodiment changes remains critical for real-world deployment.

In addition, we observed from the real-robot evaluation in the Cap Pen task that the TSAIL-HRL team consistently demonstrated strong task execution trends throughout the entire motion trajectory. However, minor inaccuracies in the final insertion phase led to a low score of 3.1. This highlights a limitation of the current evaluation protocol, where performance is overly weighted toward precise end-effector outcomes, with limited recognition of near-successful behaviors. As such, designing more expressive and sparsely distributed evaluation metrics to reflect task progress and manipulation difficulty is an important future direction.

Moreover, thanks to carefully designed data processing pipelines and algorithmic advances, several teams successfully leveraged large-scale datasets collected by industrial partners during the real-robot round. Nonetheless, issues such as anomalous values and inconsistent data quality arising from the data collection process remain unresolved. Therefore, establishing standardized pipelines for quality assurance, anomaly detection, and correction in large-scale robotic data collection is another key research challenge to be addressed for scalable and reliable model development.

\section{Conclusion}
The RoboTwin Dual-Arm Collaboration Challenge at CVPR 2025 attracted significant global participation, with over 400 researchers and engineers across 64 teams contributing to the competition. The community’s collaborative efforts led to the completion of this technical report, which consolidates key insights on policy modality selection, data composition, and multi-task learning for dual-arm manipulation.

In this report, we introduced the winning solutions—AnchorDP3 and SEM—and provided an in-depth analysis of the techniques behind their success. In addition, we highlighted valuable observations and lessons shared by participating teams across all competition stages, offering a broad perspective on the current capabilities and challenges in building robust and generalizable dual-arm manipulation systems.

We also identified several promising directions for future research, including long-horizon multi-stage reasoning, instruction-following capabilities, deformable object manipulation, robustness to domain shifts, and progress-aware evaluation metrics. These insights aim to guide the development of next-generation embodied intelligence systems that can operate effectively in dynamic, real-world environments.

We hope this challenge and its findings will inspire continued research toward generalizable, robust, and multimodal embodied intelligence for dual-arm robotic systems operating in both simulation and the physical world.

\section{Authors List (Competition Volunteers and Participants)}
\label{Competition Participants}

\textbf{Competition Volunteers}


Tian Nian, Weiliang Deng, Yiheng Ge, Yibin Liu, Zixuan Li, Dehui Wang, Zhixuan Liang, Haohui Xie, Rijie Zeng, Yunfei Ge, Peiqing Cong, Guannan He, Zhaoming Han, Ruocheng Yin, Jingxiang Guo, Lunkai Lin, Tianling Xu

\vspace{-0.4em}

\begin{multicols}{2}
\noindent\textbf{TSAIL-HRL} \\
Hongzhe Bi, Xuewu Lin, Tianwei Lin, Shujie Luo, Keyu Li \\

\textbf{JD-TFS} \\
Ziyan Zhao, Ke Fan, Heyang Xu, Bo Peng, Wenlong Gao, Dongjiang Li, Feng Jin, Hui Shen \\

\textbf{SHU-Robot} \\
Jinming Li, Chaowei Cui, Yu Chen, Yaxin Peng, Lingdong Zeng \\

\textbf{Digital Dynamos} \\
Wenlong Dong, Tengfei Li, Weijie Ke, Jun Chen, Erdemt Bao, Tian Lan, Tenglong Liu, Jin Yang, Huiping Zhuang \\

\textbf{Reconova-CAUC} \\
Baozhi Jia, Shuai Zhang, Zhengfeng Zou, Fangheng Guan, Tianyi Jia, Ke Zhou, Hongjiu Zhang, Yating Han, Cheng Fang \\

\textbf{Dexmal-Bot} \\
Yixian Zou, Chongyang Xu, Qinglun Zhang, Shen Cheng, Xiaohe Wang, Ping Tan, Haoqiang Fan, Shuaicheng Liu \\

\textbf{Colab Dual-Arm} \\
Jiaheng Chen, Chuxuan Huang, Chengliang Lin, Kaijun Luo, Boyu Yue, Yi Liu, Jinyu Chen, Zichang Tan \\

\textbf{VMV} \\
Liming Deng, Shuo Xu, Zijian Cai, Shilong Yin, Hao Wang, Hongshan Liu\\

\textbf{MOMODA} \\
Tianyang Li, Long Shi, Ran Xu \\

\textbf{Any-Robot} \\
Huilin Xu, Zhengquan Zhang, Congsheng Xu, Jinchang Yang, Feng Xu \\

\end{multicols}





















\clearpage
{
\small
\bibliographystyle{plain}
\bibliography{ref}
}

\end{document}